\documentclass[a4paper,conference]{IEEEtran}

\usepackage[nottoc]{tocbibind}
\usepackage{graphicx}
\usepackage{comment}
\usepackage{amsmath,amssymb} 
\usepackage{color}
\usepackage{algpseudocode}
\usepackage{algorithm}
\usepackage{amsfonts}
\usepackage{multirow}
\usepackage{tablefootnote}
\usepackage{adjustbox}
\usepackage{threeparttable}

\algnewcommand{\LineComment}[1]{\State \(\triangleright\) #1}
\algnewcommand\algorithmicforeach{\textbf{for each}}
\algdef{S}[FOR]{ForEach}[1]{\algorithmicforeach\ #1\ \algorithmicdo}

\begin{document}
%
\title{Learning to Segment Dynamic Objects\\ using SLAM Outliers}

\author{\IEEEauthorblockN{Adrian Bojko\IEEEauthorrefmark{1},
Romain Dupont\IEEEauthorrefmark{2}, Mohamed Tamaazousti\IEEEauthorrefmark{3} and
Hervé Le Borgne\IEEEauthorrefmark{4}}
\IEEEauthorblockA{Université Paris-Saclay, CEA, List, F-91120, Palaiseau, France\\
Email: \IEEEauthorrefmark{1}adrian.bojko@cea.fr,
\IEEEauthorrefmark{2}romain.dupont@cea.fr,
\IEEEauthorrefmark{3}mohamed.tamaazousti@cea.fr,
\IEEEauthorrefmark{4}herve.le-borgne@cea.fr}}

\maketitle

\begin{abstract}
We present a method to automatically learn to segment dynamic objects using SLAM outliers. It requires only one monocular sequence per dynamic object for training and consists in localizing dynamic objects using SLAM outliers, creating their masks, and using these masks to train a semantic segmentation network. We integrate the trained network in ORB-SLAM 2 and LDSO. At runtime we remove features on dynamic objects, making the SLAM unaffected by them. We also propose a new stereo dataset and new metrics to evaluate SLAM robustness. Our dataset includes consensus inversions, i.e., situations where the SLAM uses more features on dynamic objects that on the static background. Consensus inversions are challenging for SLAM as they may cause major SLAM failures. Our approach performs better than the State-of-the-Art on the TUM RGB-D dataset in monocular mode and on our dataset in both monocular and stereo modes.
\end{abstract}

\IEEEpeerreviewmaketitle

\section{Introduction}
Simultaneous Localization and Mapping (SLAM) algorithms \cite{mouragnon2006real, klein_parallel_2007} are frequently used in autonomous vehicles \cite{rosinol_ultimate_2018}, augmented reality \cite{besbes2012interactive, gay2012mobile, tamaazousti2016constrained} and robotics \cite{cheng_accurate_2018}. A SLAM localizes the camera in a static environment while reconstructing it \cite{cadena_past_2016}. Visual SLAMs rely on image features and assume that the camera moves in a static environment \cite{hartley_2003} (static world assumption), although it is rarely met in practice.

An extreme case where the static world assumption is not true is the motion consensus inversion, which we define as a situation where the SLAM relies more on dynamic objects than on the static environment. This may happen when an object close to the camera moves. The SLAM effectively uses a frame of reference that is not the ground thus fails. Some works give examples of such situations~\cite{schorghuber_slamantic_2019,barnes_driven_2018} without further studying the problem. In this article, we specifically address this issue that is of high interest in practice.

Dynamic SLAMs aim to reduce the impact of dynamic objects by explicitly processing them. Most Dynamic SLAMs are constructed by improving an existing SLAM through different approaches. \textit{Geometry-based approaches} detect motion at runtime~\cite{cheng_improving_2019} using geometrical algorithms, e.g., optical flow. These approaches compute the dominant motion of the image and may fail if there are motion consensus inversions. 

\textit{Learning-based approaches} use semantic masks \cite{kaneko_mask-slam:_2018}. Classes are defined at annotation time (before training), and any class that is not annotated is not recognized at runtime.
\textit{Hybrid approaches} try to compensate the limits of one approach with the advantages of the other. The first type combines geometrical and learned approaches at runtime \cite{bescos_dynaslam:_2018}, but our experiments show that it does not prevent consensus inversions. The second type uses geometry during training and semantic segmentation at runtime. An example is \cite{barnes_driven_2018}: although their approach reduces the risk of consensus inversion, it needs training sequences recording the same location at different times, does not detect objects that do not move between training sequences and requires a full stereo camera + LIDAR setup.

\begin{figure}[t]
    \centering
    \includegraphics[width=0.485\textwidth]{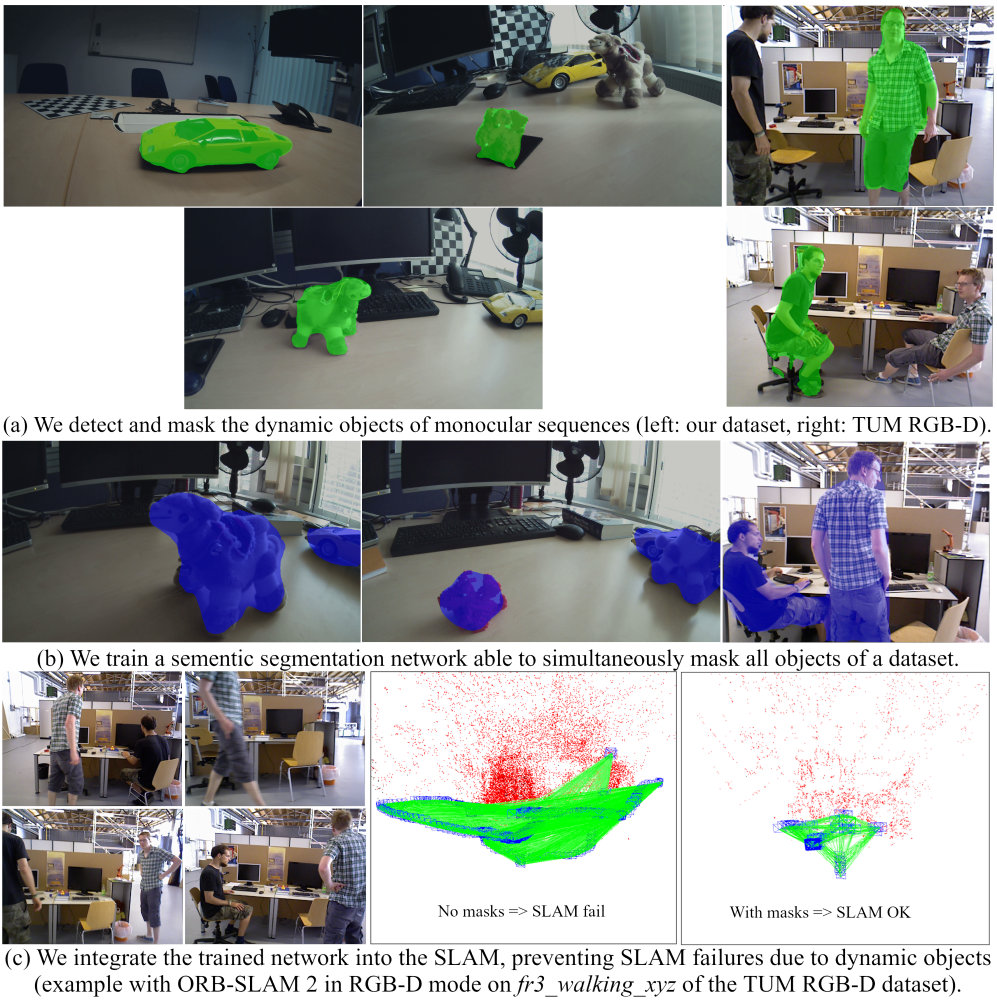} 
    \caption{Steps of our method. We only need one monocular sequence per dynamic object to make the SLAM robust and prevent major failures due to dynamic objects. We train one network for our Consensus Inversion dataset and one for TUM RGB-D.}
    \label{fig:visual_abstract}
\end{figure}

Similarly to \cite{barnes_driven_2018}, we use geometrical information (inliers and outliers) to localize dynamic objects and construct their masks, which we use to train a network. We integrate the trained network in an existing SLAM to make it Dynamic. However, our approach only needs one monocular sequence per dynamic object, while \cite{barnes_driven_2018} requires at least two and up to eight in practice. Moreover, contrary to~\cite{barnes_driven_2018}, we do not need stereo nor depth information, making our approach easier to implement and put into practice. 

We use our method to make ORB-SLAM 2 \cite{mur-artal_orb-slam2:_2017} (Monocular, Stereo and RGB-D) and LDSO \cite{gao_ldso_2018} robust to dynamic objects without any manual annotations. In addition to our method, we created the dataset \emph{Consensus Inversion} and created two new metrics, Penalized ATE RMSE and Success Rate, to quantify the robustness of Dynamic SLAMs.

\section{Related Work}
Given a geometric SLAM algorithm, making it robust to dynamic objects is relevant to real-world applications since it generalizes the SLAM from static to dynamic environments. Keeping the original, non-learned modules of an existing SLAM is valuable as it limits generalization issues encountered by learned algorithms, e.g., end-to-end SLAMs \cite{wang_end--end_2018}. For these reasons, we focus on Dynamic SLAMs that keep the original SLAM modules.

Such Dynamic SLAMs include a dynamic object masking module in their pipeline. Most have the masking module placed right after feature computation: they either remove features on dynamic objects or process them separately. Rarely, some Dynamic SLAMs start by masking dynamic objects then detect features only in unmasked areas of the image \cite{kaneko_mask-slam:_2018}.

\subsection{Geometry-based approaches}
Geometry-based approaches do not use any kind of machine learning. \cite{saputra_visual_2018} surveys many non-learned Dynamic SLAMs and concludes with "handling missing, noisy, and outlier data remains a future challenge for most of the discussed techniques [...] Most techniques also have difficulty in dealing with degenerate and dependent motion.". Optical flow approaches \cite{cheng_accurate_2018} compute pixel displacements between frames but may not work if dynamic objects occupy most of the scene or have an erratic motion. Depth maps approaches \cite{sun_improving_2017} use the additional depth information to identify salient objects but are limited by sensor range and resolution. Clustering/background-foreground approaches \cite{li_rgb-d_2017,sun_motion_2018} identify dynamic objects by grouping and assigning probabilities to points with similar motions but have high computational costs and do not work well with noisy or degenerate motions \cite{saputra_visual_2018}. Our experiments show that geometrical approaches do not completely prevent consensus inversions.

\subsection{Learning-based approaches that use semantic masks}
Learning-based approaches that use semantic masks \cite{kaneko_mask-slam:_2018,wang_unified_2019} remove features on dynamic objects at runtime according to their class (e.g. people and cars are usually masked) but are limited by the availability and scope of training data, not working with unknown objects \cite{yu_ds-slam:_2018}.

\subsection{Hybrid approaches}
Hybrid approaches are divided in two types.

The first type combines geometrical and learned algorithms at runtime \cite{bescos_dynaslam:_2018,brasch_semantic_2018,li_stereo_2018,schorghuber_slamantic_2019,yu_ds-slam:_2018}, leveraging strong points while compensating weaknesses (as optical flow + semantics or depth map + semantics). However, there are still failure points common to the underlying algorithms, e.g., an object of unknown class causing a consensus inversion.

The second type uses geometrical approaches during training and learned approaches at runtime \cite{barnes_driven_2018}. During training, geometrical approaches are used to generate the masks of dynamic objects, which are then learned. At runtime it is the same as learning-based approaches. \cite{barnes_driven_2018} learns the appearance of dynamic objects by comparing 3D reconstructions over time but requires several traversals of the same environment at different times and a full stereo camera + LIDAR setup during training. Our approach is like \cite{barnes_driven_2018} but we require only one monocular sequence per object, without needing additional traversals or sensors.

\section{Learning to Segment Dynamic Objects}
\subsection{Overview}
\begin{figure*}
    \centering
    \includegraphics[width=0.7\textwidth]{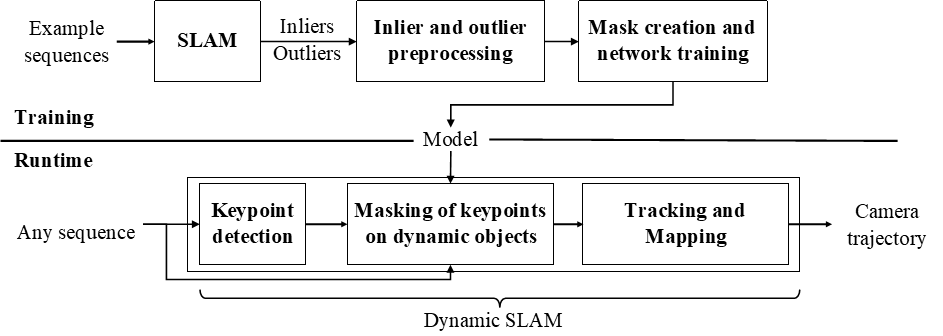} 
    \caption{Overview of our approach. We collect inliers and outliers from example sequences and use them to create masks of dynamic objects. We train a semantic segmentation network with the created masks and integrate it in the SLAM after the keypoint detection step. At runtime we infer the masks of any sequence and remove all features on dynamic objects.}
    \label{fig:overview}
\end{figure*}

We outline in this section our approach, illustrated in Fig. \ref{fig:overview}, to generalize a feature-based SLAM into a Dynamic SLAM. Our goal is to protect the SLAM against the negative effects of dynamic objects in a given environment. We achieve this by training a segmentation network with two classes: \emph{static} and \emph{dynamic}, the latter being masked during the execution of the SLAM. Having detailed classes as \emph{car} or \emph{person} is unneeded since we always mask dynamic objects. We show that unconditional masking is more robust than geometry, which fails under consensus inversions. Among mask-based approaches, ours is the only one with automatic annotation and very low data requirements (one sequence per object).

SLAMs usually reject non-static features with methods as RANSAC \cite{mur-artal_orb-slam2:_2017}. We make the hypothesis that, if there is no motion consensus inversion, the sudden apparition of dense clusters of outliers characterizes violations of the static world assumption by dynamic objects. Thus, if dense clusters of outliers suddenly replace inliers it means that the inliers that became outliers were in fact dynamic features, indicating that there is a whole \emph{object} violating the static world assumption rather than just isolated features. 

Given a SLAM, we define example sequences as sequences that respect our hypothesis, i.e., make the SLAM generate clusters of outliers on dynamic objects when they move. In practice, example sequences are sequences in which dynamic objects are reconstructed by the SLAM and do not cause motion consensus inversions, e.g., a sitting person that stands up a couple meters away from the camera.

The steps of our approach are:

\begin{enumerate}
    \item \textbf{Outlier and inlier preprocessing}: we use the SLAM to generate outliers and inliers. For non-deterministic SLAMs, we add a filtering step.
    \item \textbf{Mask creation and network training}: we use the inliers and outliers to create the masks of dynamic objects. Then we use the masks to train a network.
    \item \textbf{SLAM Integration}: we integrate the trained network right after the feature detection step of the SLAM.
\end{enumerate}

\subsection{Outlier and inlier preprocessing}
Once initialized, a feature-based SLAM algorithm computes camera poses for each frame in roughly three steps: 
\begin{enumerate}
    \item 2D keypoint detection.
    \item 2D-3D matching between detected keypoints and known 3D map points + triangulation of new 3D map points.
    \item Bundle adjustment: robust optimization of 2D-3D matches and camera poses.
\end{enumerate}

We save the coordinates of outliers and inliers of each frame right after the bundle adjustment: outliers are keypoints whose 2D-3D match was rejected, and inliers those whose 2D-3D match was not rejected. SLAMs may be non-deterministic due to multithreading or random functions (e.g,. RANSAC). So, we save inliers and outliers coordinates over several runs and merge them, filtering rarely observed coordinates as they tend to create spurious clusters when merged.

\subsection{Mask creation and network training}
This step is divided in mask creation network training.

\subsubsection{Mask creation} this step consists in localizing dynamic objects with sliding windows, refining the windows into masks, and propagating the masks to the whole sequence.

\textbf{Localizing dynamic objects with sliding windows}: on every image of every sequence we use rectangular sliding windows of different sizes, at a fixed stride, to evaluate how the inlier/outlier ratio changes.  

Let $w$ be a window on image $p$ and $w'$ its corresponding window on image $p'$. Then the outlier score $S$ is:
\begin{equation}
    S = \left(\frac{\text{outlier density of $w$}}{\text{inlier density of $w$}}\right) / \left({\frac{\text{outlier density of $w'$}}{\text{inlier density of $w'$}}}\right)
\end{equation}

When a mapped object moves, many inliers become outliers, making the ratio $S$ drop between consecutive frames. We consider that $w$ contains a dynamic object if $S$ is less than a threshold $S_{max}$, set by the user. 

We roughly compensate camera motion with the homography $H=K.dR_{p,p'}.K^{-1}$ where $K$ is the camera intrinsic matrix and $dR_{p,p'}$ is the relative rotation between $w$ and $w'$. We apply $H$ on window $w'$ to have both $w$ and $w'$ match the same physical location. This approximation is easy to compute using a trajectory generated by the SLAM and was accurate enough in our experiments.

\textbf{Refining sliding windows into single masks}: we merge all bounding boxes that overlap on the same image. Then we project each merged bounding box on the past and future $k$ frames and create image sequences with the content of these projected bounding boxes. The created sequences are a perfect fit for Unsupervised Video Object Segmentation (UVOS, methods that automatically segment salient/dynamic objects in videos) as there is no ambiguity on which object to segment. For each created sequence we apply the author's implementation of COSNet~ \cite{lu_see_2019} (a State-of-the-Art UVOS network) on the central images, thus masking the dynamic objects inside the sliding windows.

\textbf{Propagating single masks}: now that we have a single accurate binary mask for every dynamic object, we can propagate them to past and future frames using semi-supervised video object segmentation. These methods track dynamic objects in videos but require very accurate initial guesses which we have thanks to the previous step. We apply the author's implementation of SiamMask \cite{wang_fast_2019} (a State-of-the-Art network that is both lightweight and class-agnostic) towards past and future frames. The result is a set of binary masks, covering the whole sequence, for each dynamic object.

\subsubsection{Network training} Our goal is to train a semantic network able to mask all dynamic objects simultaneously. First, we train one instance of DeepLabv3+\footnote{Source: https://github.com/srihari-humbarwadi/person\_segmentation\_tf2.0}~\cite{chen_encoder-decoder_2018} (a State-of-the-Art semantic segmentation network) for every set of masks generated at the previous step. Then, we infer masks for every sequence and for every trained network. We superimpose the masks inferred on the same sequence and use all superimposed masks to train a final instance of DeepLabv3+. The computed model can be used to mask all dynamic objects of all sequences simultaneously.

\subsection{SLAM Integration}
We integrate the final DeepLabv3+ trained network after the feature detection module in the SLAM. We infer the mask of dynamic objects from the current image, then filter all features whose coordinates are on masked areas.

\section{Consensus Inversion Dataset and new metrics} \label{sec:consensus_inversion_dataset}
\begin{figure}
    \centering
    \includegraphics[width=0.5\textwidth]{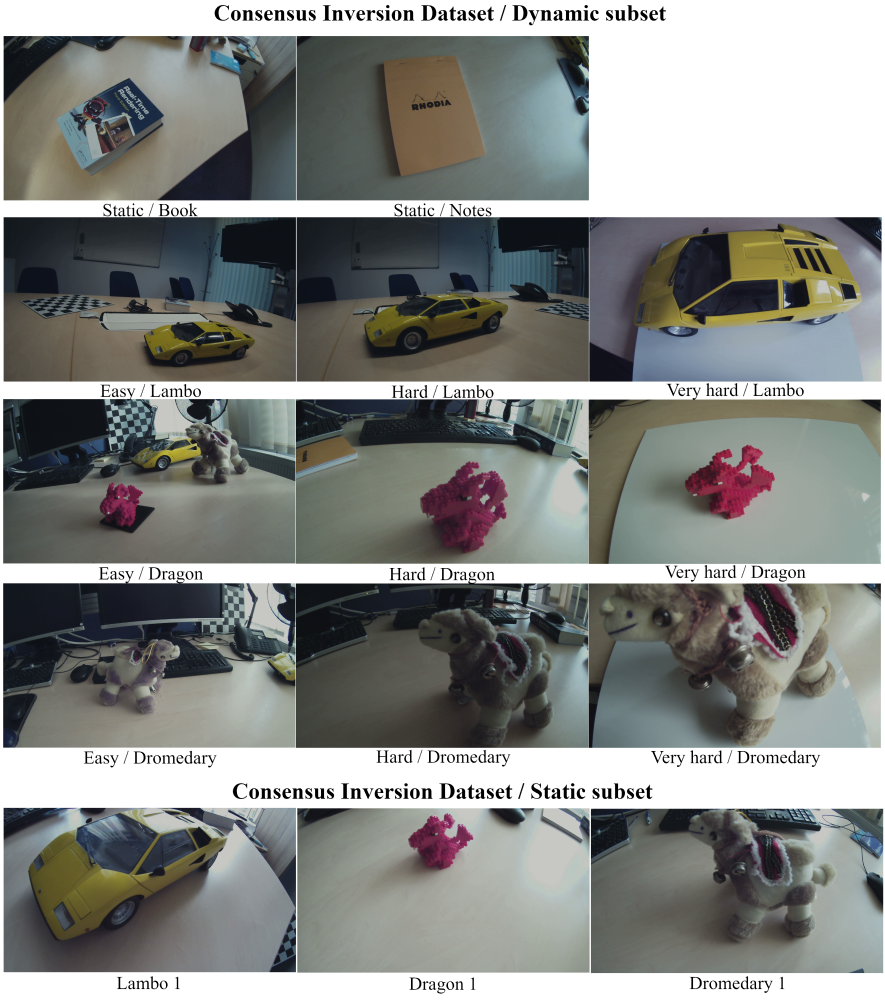} 
    \caption{Miniatures of our dataset Consensus Inversion. The camera moves in the Dynamic subset and stays static in the Static subset.}
    \label{fig:miniatures}
\end{figure}

To the best of our knowledge, there is no dataset nor appropriate metrics to test the robustness of SLAMs with respect to motion consensus inversions. Hence, we created the \emph{Consensus Inversion} dataset, illustrated in Fig. \ref{fig:miniatures}, made of two subsets: Dynamic and Static. We used a MYNT EYE D1000-120 stereo camera at 1280x720 / 30Hz. IMU data is recorded but not used. All sequences are about 500 to 1000 images long.

\subsection{Consensus Inversion / Dynamic} \label{sec:consensus_inversion_dynamic}
The camera is dynamic in the first subset. It includes two sequences with static objects (book/notes) and twelve with dynamic objects (dragon/dromedary/car, each object moves in three sequences). This subset covers different situations:
\begin{itemize}
    \item \emph{Static}: a static object close to the camera. Tests oversegmentation, i.e., if the SLAM masks a static object and fails for this reason (all image features are masked).
    \item \emph{Easy}: an object moves but does not cause consensus inversion. Tests standard SLAM robustness.
    \item \emph{Hard}: an object moves and causes motion consensus inversion. Tests SLAM robustness to consensus inversions.
    \item \emph{Very hard}: an object moves rigidly with the camera while very close to it. Tests the robustness to consensus inversion when detecting object motion is extremely difficult. 
\end{itemize}

We computed the ground truth and the ground truth tracking rate (\% of tracked frames) using ORB-SLAM 2 stereo improved with our method and with the early stopping of bundle adjustment removed. We also computed the ground truth tracking rate in monocular mode.

\subsection{Consensus Inversion / Static}
The camera is static in the second subset. An object close to the camera starts moving and causes a consensus inversion: the SLAM must, however, not compute any motion. We made five sequences per dynamic object.

\subsection{Metrics: Penalized ATE RMSE and Success Rate} \label{sec:new_metrics}
The standard metrics to test SLAMs are the ATE RMSE (Absolute Trajectory Error) and the tracking rate (\% of tracked frames). However, if tracking rates are too different the comparison of ATE RMSEs is biased: a SLAM that stops early might skip tricky parts of the sequence. Hence, we defined the Penalized ATE RMSE and Success Rate.

We consider that an ATE RMSE is invalid if: 1) it is unknown (e.g. when using reported results) or 2) the tracking rate is lower than $r_{gt}-\Delta_{r,max}$, where $\Delta_{r,max}$ is a fixed threshold and $r_{gt}$ is the ground truth tracking rate (i.e. the expected \% of tracked frames). 

The Penalized ATE RMSE is computed in relation to other SLAMs in case of invalidation. With $\tau$ the penalty factor and $L$ the set of all valid ATE RMSEs computed by other SLAMs on the tested sequence, we define the Penalized ATE RMSE in Eq. \ref{alg:penalized}:

\begin{equation} \label{alg:penalized}
\resizebox{0.49 \textwidth}{!} 
{$\text{Penalized ATE RMSE}
  \begin{cases}
    \max(L).(1+\tau), & \text{if ATE RMSE is invalid}.\\
    \text{ATE RMSE}, & \text{otherwise}.
  \end{cases}$}
\end{equation}

Conversely, we consider that a SLAM successfully processes a sequence if all these conditions are respected: 1)~The ATE RMSE is known 2) The ATE RMSE is below a fixed threshold $l_{max}$ 3) The tracking rate is at least $r_{gt}-\Delta_{r,max}$. The Success Rate of a SLAM on a dataset is the ratio of sequences that the SLAM successfully processes.

\section{Experiments} \label{sec:experiments}
\subsection{Experimental setup}
\subsubsection{Datasets and parameters} we evaluate our method on the TUM RGB-D dataset \cite{sturm_benchmark_2012} (dynamic sequences) and on the Consensus Inversion dataset. The purpose of the TUM RGB-D dataset is the evaluation of RGB-D SLAM systems; it was recorded using a Microsoft Kinect and the ground truth camera poses was obtained from a motion capture system. The sequences contain color and depth images at 640x480/30Hz. The dynamic sequences are a set of eight sequences that record people moving.

We use ORB-SLAM 2 \cite{mur-artal_orb-slam2:_2017} as the core SLAM algorithm for the main experiments and LDSO \cite{gao_ldso_2018} specifically to test the extension to a Direct SLAM. We set the feature number to 3000 and use default/author settings otherwise. We use our method to train one network on the TUM RGB-D dataset (we use the sequences fr3\_sitting\_static and fr3\_walking\_static) and one on the Consensus Inversion dataset (we use the \emph{Easy} sequences of the Dynamic subset).

We empirically found suitable parameters: sliding windows of size 100x100 / 200x200 / 300x300 / 400x400 with a stride of 50px, a difference of 3 images to compute outliers scores, an interval $k=15$ images (1s at 30Hz) for the unsupervised VOS and a max outlier score $S_{max} = 0.15$ to determine if a window contains a dynamic object.

\begin{figure}
    \centering
    \includegraphics[width=0.49\textwidth]{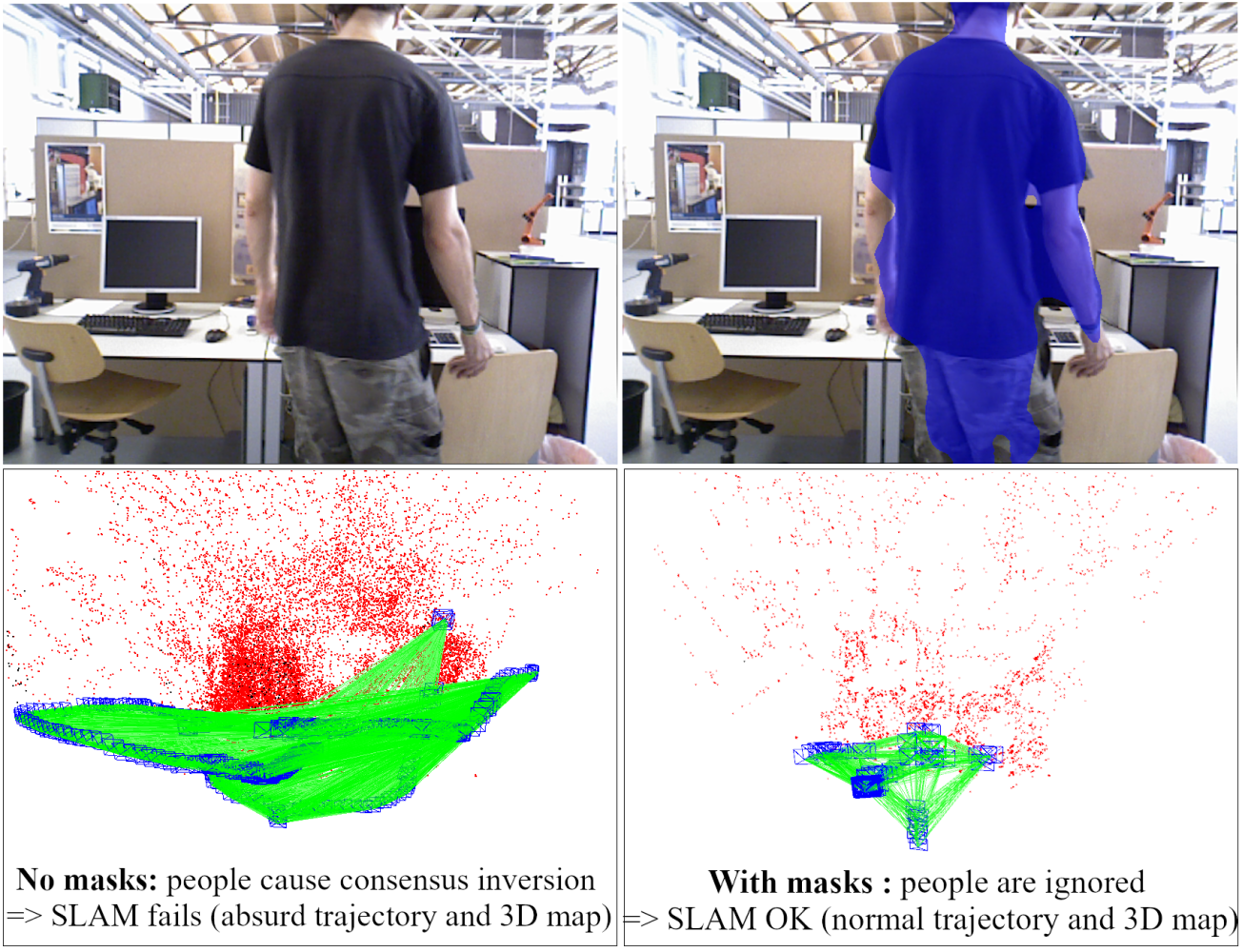} 
    \caption{Consensus inversion in fr3\_walking\_xyz (TUM RGB-D). Camera pose in blue and 3D map in red. Left: no masks, the camera trajectory is nonsensical as the SLAM uses features on moving people. Right: we apply masks using our method, the SLAM trajectory is coherent with the real motion.}
    \label{fig:tum_inversion}
\end{figure}

\begin{figure}
    \centering
    \includegraphics[width=0.49\textwidth]{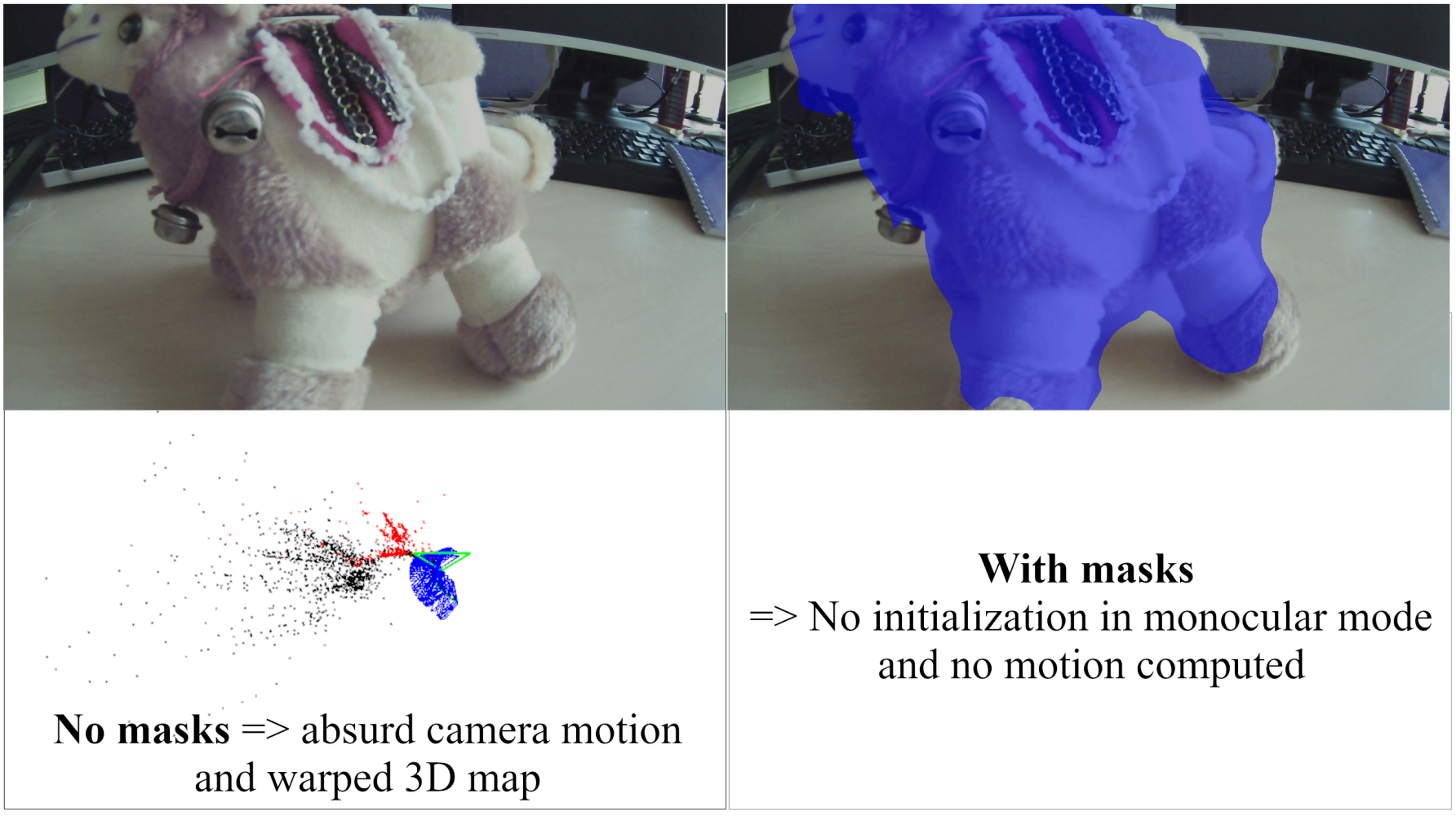} 
    \caption{Example of monocular false start. The SLAM cannot initialize as the camera is perfectly static. Yet if the object is not masked the SLAM generates absurd trajectories and 3D maps. Masking dynamic objects prevents such situations.}
    \label{fig:static_cam}
\end{figure}

\begin{figure}
    \centering
    \includegraphics[width=0.49\textwidth]{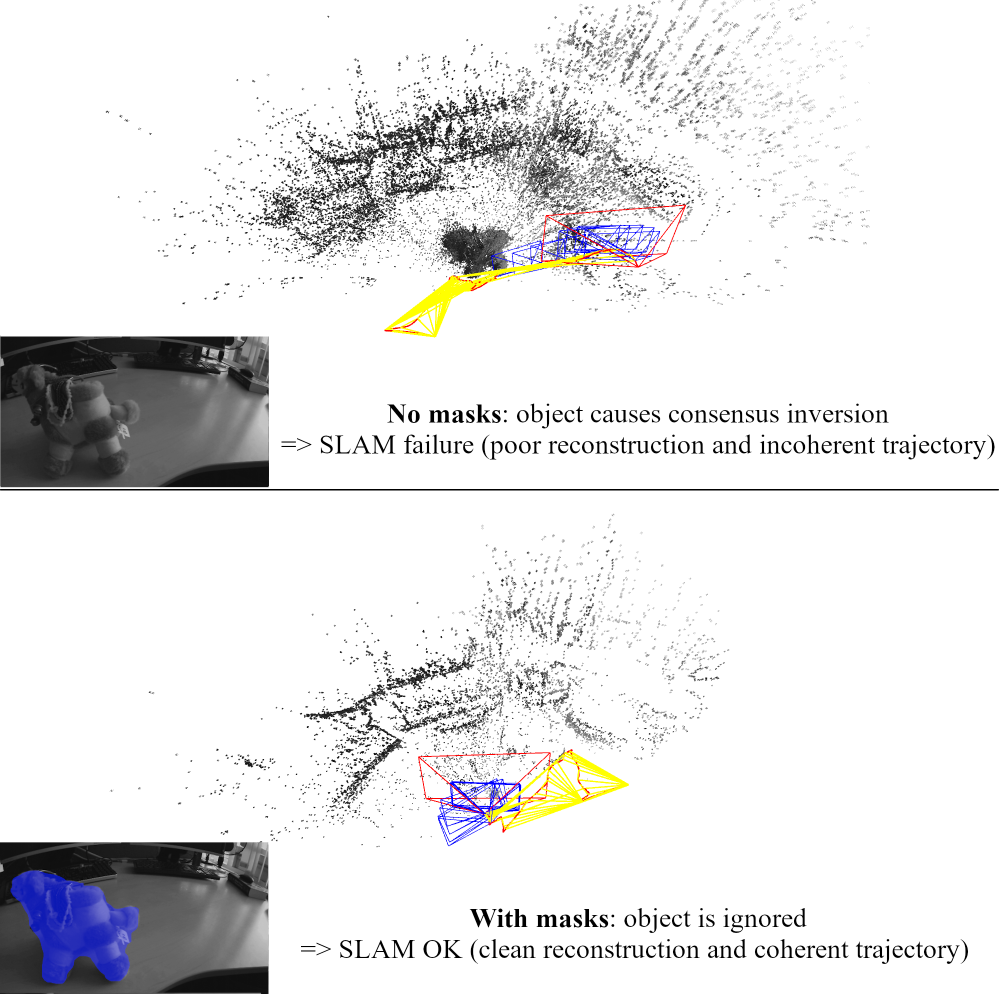} 
    \caption{We apply our method on LDSO and test it on a hard sequence. Without masks the SLAM fails, with masks it ignores the object and works correctly.}
    \label{fig:ldso_inversion}
\end{figure}

\subsubsection{Metrics}
To consider non-deterministic behaviors, we compute the median of the ATE RMSE of the keyframe trajectory with a Sim(3) alignment and the tracking rate over 10 executions for every sequence. We then compute the Penalized ATE RMSE and Success Rate (section \ref{sec:new_metrics}) with penalty coefficient $\tau = 10\%$, max acceptable tracking rate decrease $\Delta_{r,max} = 15\%$ and max acceptable ATE RMSE $l_{max} = 10cm$.

\subsection{Results}

\begin{table*}
\Huge
    \centering
    \resizebox{\textwidth}{!}{
        \renewcommand{\arraystretch}{1.5}
        \begin{tabular}{|l|c|c|c|c|c|c|c|c|c|c|c|}
        
        \cline{2-12}
    \multicolumn{1}{c|}{} & \multicolumn{5}{|c|}{\textbf{State-of-the-Art}} & \multicolumn{6}{|c|}{\textbf{ORB-SLAM 2} \cite{mur-artal_orb-slam2:_2017} + $\cdots$}  \\ \cline{2-12} 
    \multicolumn{1}{c|}{} & \multirow{2}{*}{{L-K}\textsuperscript{\ref{ft:lk}}\cite{cheng_accurate_2018}} & \multirow{2}{*}{{Dyna}\textsuperscript{\ref{ft:dynaslam}}\cite{bescos_dynaslam:_2018}} & \multirow{2}{*}{{ST}\textsuperscript{\ref{ft:slamantic}}\cite{schorghuber_slamantic_2019}} & \multirow{2}{*}{{Uni}\textsuperscript{\ref{ft:refer}}\cite{wang_unified_2019}} & \multirow{2}{*}{{DS}\textsuperscript{\ref{ft:refer}}\cite{yu_ds-slam:_2018}} & \multicolumn{5}{|c|}{\textbf{Segmentation baselines}} & \multirow{2}{*}{\textbf{Our seg.}} \\ \cline{1-1}
        \textbf{Test dataset} & & & & & & {No seg.} & {Mask R-CNN\cite{he_mask_2017}} & {PWC-Net\cite{sun_pwc-net_2018}} & {RVOS\cite{ventura_rvos_2019}} & {COSNet\cite{lu_see_2019}} & \\ \hline
        Consensus Inversion - Mono & 0,0547 & 0,0693 & 0,0692 & N/A & N/A & 0,0860 & 0,0760 & 0,0237 & 0,0144 & 0,0297 & \textbf{0,0089} \\ \hline
        Consensus Inversion - Stereo & N/A & 0,0627 & 0,0699 & N/A & N/A & 0,0756 & 0,0630 & 0,0803 & 0,0116 & 0,0148 & \textbf{0,0094} \\ \hline
        TUM RGB-D - Mono & 0,0892 & 0,1108 & 0,1101 & N/A & N/A & 0,0252 & 0,0235 & 0,0335 & 0,0331 & 0,0267 & \textbf{0,0222} \\ \hline
        TUM RGB-D - RGB-D & N/A & 0,0206 & 0,0173 & 0,0190 & 0,0802 & 0,1077 & \textbf{0,0172} & 0,0790 & 0,0218 & 0,0245 & 0,0185 \\ \hline    
    \end{tabular}}
    \vspace*{0mm}
    
    \caption{\small Average Penalized ATE RMSE (m) of the State-of-the-Art and baselines on Consensus Inversion/Dynamic and TUM RGB-D/Dynamic datasets. N/A indicates that the SLAM mode is not supported.}
    \label{tab:ate_sota}
\end{table*}

\begin{table*}
\huge
    \centering
    \resizebox{\textwidth}{!}{
    \renewcommand{\arraystretch}{1.5}

    \begin{tabular}{|l|c|c|c|c|c|c|c|c|c|c|c|}
    \cline{2-12}
    \multicolumn{1}{c|}{} & \multicolumn{5}{|c|}{\textbf{State-of-the-Art}} & \multicolumn{6}{|c|}{\textbf{ORB-SLAM 2} \cite{mur-artal_orb-slam2:_2017} + $\cdots$}  \\\cline{2-12} 
    \multicolumn{1}{c|}{} & \multirow{2}{*}{{L-K}\textsuperscript{\ref{ft:lk}}\cite{cheng_accurate_2018}} & \multirow{2}{*}{{Dyna}\textsuperscript{\ref{ft:dynaslam}}\cite{bescos_dynaslam:_2018}} & \multirow{2}{*}{{ST}\textsuperscript{\ref{ft:slamantic}}\cite{schorghuber_slamantic_2019}} & \multirow{2}{*}{{Uni}\textsuperscript{\ref{ft:refer}}\cite{wang_unified_2019}} & \multirow{2}{*}{{DS}\textsuperscript{\ref{ft:refer}}\cite{yu_ds-slam:_2018}} & \multicolumn{5}{|c|}{\textbf{Segmentation baselines}} & \multirow{2}{*}{\textbf{Our seg.}} \\ \cline{1-1}
    \textbf{Test dataset} & & & & & & {No seg.} & {Mask R-CNN\cite{he_mask_2017}} & {PWC-Net\cite{sun_pwc-net_2018}} & {RVOS\cite{ventura_rvos_2019}} & {COSNet\cite{lu_see_2019}} & \\ \hline
    Consensus Inversion - Mono & 54,5\% & 63,6\% & 63,6\% & N/A & N/A & 45,5\% & 54,5\% & 72,7\% & 72,7\% & 72,7\% & \textbf{100,0\%} \\ \hline
    Consensus Inversion - Stereo & N/A & 72,7\% & 63,6\% & N/A & N/A & 63,6\% & 63,6\% & 63,6\% & 81,8\% & 81,8\% & \textbf{100,0\%} \\ \hline
    TUM RGB-D - Mono & 50,0\% & 62,5\% & 62,5\% & N/A & N/A & 87,5\% & 87,5\% & 62,5\% & 62,5\% & \textbf{100,0\%} & \textbf{100,0\%} \\ \hline
    TUM RGB-D - RGB-D & N/A & \textbf{100,0\%} & \textbf{100,0\%} & \textbf{100,0\%} & 87,5\% & 62,5\% & \textbf{100,0\%} & 62,5\% & \textbf{100,0\%} & \textbf{100,0\%} & \textbf{100,0\%} \\ \hline    
    \end{tabular}}
    \vspace*{0mm}
    \caption{\small Success Rate (\%) of the State-of-the-Art and baselines on Consensus Inversion/Dynamic and TUM RGB-D/Dynamic datasets. N/A indicates that the SLAM mode is not supported.}
    \label{tab:success_sota}
\end{table*}

\begin{table*}
    \centering
    \renewcommand{\arraystretch}{1.5}
    \begin{tabular}{|c|c|c|c|c|c|c|c|c|c|c|}
    \hline
    \multicolumn{3}{|c|}{\textbf{State-of-the-Art}} & \multicolumn{6}{|c|}{\textbf{ORB-SLAM 2} \cite{mur-artal_orb-slam2:_2017} + $\cdots$}  \\ \hline
    \multirow{2}{*}{{L-K}\textsuperscript{\ref{ft:lk}}\cite{cheng_accurate_2018}} & \multirow{2}{*}{{Dyna}\textsuperscript{\ref{ft:dynaslam}}\cite{bescos_dynaslam:_2018}} & \multirow{2}{*}{{ST}\textsuperscript{\ref{ft:slamantic}}\cite{schorghuber_slamantic_2019}} & \multicolumn{5}{|c|}{\textbf{Segmentation baselines}} & \multirow{2}{*}{\textbf{Our seg.}} \\
    & & & {No seg.} & {Mask R-CNN\cite{he_mask_2017}} & {PWC-Net\cite{sun_pwc-net_2018}} & {RVOS\cite{ventura_rvos_2019}} & {COSNet\cite{lu_see_2019}} & \\ \hline
    53,3\% & 60,0\% & 60,0\% & 60,0\% & 60,0\% & 66,7\% & 86,7\% & 80,0\% & \textbf{100,0\%} \\ \hline    \end{tabular}
    \vspace*{1mm}
    \caption{\small Evaluation on Consensus Inversion/Static dataset. We report the ratio of sequences that do not cause initialization fails (false starts).}
    \label{tab:sucess_static_cam}
\end{table*}

\subsubsection{Comparison with the State-of-the-Art} dynamic objects, especially when causing consensus inversions, may cause early SLAM failure and decrease in the tracking rate of SLAMs, making the comparison of ATE RMSEs biased. To take both the trajectory error and the tracking rate into account we use our new metrics: the Penalized ATE RMSE and the Success Rate. The Penalized ATE RMSE integrates failures, so it is directly comparable between SLAMs, and a higher Success Rate expresses that a SLAM is less affected by dynamic objects. We evaluate the methods on the TUM RGB-D and Consensus Inversion datasets.

Table \ref{tab:ate_sota} (cols. 2-6) shows the Penalized ATE RMSE of the State-of-the-Art. Our method performed better than others on our dataset (all modes) and on TUM RGB-D in monocular mode. It is in third place on TUM RGB-D in RGB-D mode. 

On TUM RGB-D in monocular mode the results of L-K\footnote{We implemented a simplified version of \cite{cheng_accurate_2018} (which uses the Lucas-Kanade optical flow): we warp frames with an homography and we set the optical flow displacement threshold to 2px.\label{ft:lk}}\cite{cheng_accurate_2018}, DynaSLAM\footnote{DynaSLAM randomly crashed in RGB-D mode on our system. We refer to the original results in this mode.\label{ft:dynaslam}} \cite{bescos_dynaslam:_2018} and SLAMANTIC\footnote{The publicly available code of SLAMANTIC does not support monocular mode so we adapted the available stereo code.\label{ft:slamantic}} \cite{schorghuber_slamantic_2019} are explained by the harsh penalty we give to early failures (which happened to the three of them). The original ORB-SLAM 2 already performs well so removing dynamic objects is not really necessary. All Dynamic SLAMs\footnote{We report the results of DS-SLAM and Unified.\label{ft:refer}} performed well in RGB-D including our method. We reached the standard of other Dynamic SLAMs that rely on manually annotated networks. Fig. \ref{fig:tum_inversion} illustrates how the SLAM can output nonsense in presence of consensus inversions (motions that do not exist) and that we prevent it. 

On our dataset other Dynamic SLAMs performed poorly: they failed in hard / very hard sequences when the object was of an unknown class, e.g., dragon or dromedary. Even SLAMANTIC, that tries not to segment mobile objects that are not moving (e.g. a parked car) also failed in the very hard sequences as the objects are static during most of the sequence.

Table \ref{tab:success_sota} (cols. 2-6) shows the Success Rate of the State-of-the-Art. Our approach has the best performance in all cases. The results are coherent with the Penalized ATE RMSE: a higher Success Rates correspond to a lower Penalized ATE RMSE. Except for TUM RGB-D in RGB-D mode (success rate $\geq 85\%$), all results are below 75\%. The results in monocular mode show that it is more difficult to handle monocular dynamic sequences with geometrical approaches, possibly due to the arbitrary scale of the SLAM. The  success rates on our dataset (about 60\%) shows that both geometrical approaches and hybrid ones combining geometrical/semantic approaches at runtime fail if there are consensus inversions caused by unknown objects. 

Our results prove that regarding the robustness to dynamic objects: 1) semantic networks should be fine-tuned to the dynamic objects of the working environment 2) geometrical approaches are unreliable under consensus inversions 3) Hybrid/combined approaches are unreliable under consensus inversion caused by unknown objects.

\subsubsection{Comparison with baselines} Unsupervised Video Object Segmentation (UVOS) networks and semantic segmentation networks may appear as trivial solutions to make a SLAM Dynamic as their integration is straightforward. 

To test this aspect, we integrated Mask R-CNN \cite{he_mask_2017} (we filter the same semantic classes as DynaSLAM), RVOS \cite{ventura_rvos_2019} (in zero-shot mode) and COSNet \cite{lu_see_2019} (we only use past frames for inference) in ORB-SLAM 2. We also test a very simple optical flow solution with PWC-Net \cite{sun_pwc-net_2018}, a learned optical flow network, by masking the area of the image with the 50\% most intense optical flow.

\begin{figure}
    \centering
    \includegraphics[width=0.49\textwidth]{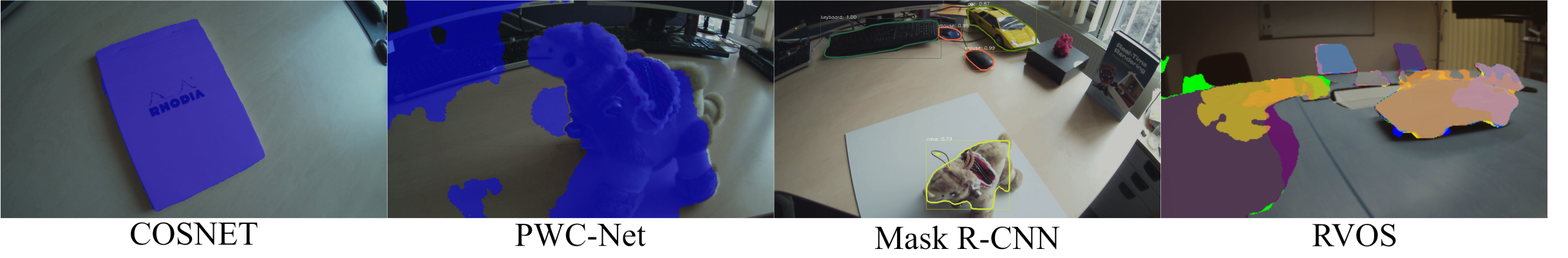} 
    \caption{Failure cases of baselines methods (ORB-SLAM 2 + existing network). Mask R-CNN ignores the dragon and considers the dromedary a cake. Other methods segment static objects.}
    \label{fig:uvos}
\end{figure}

Table \ref{tab:ate_sota} (cols. 7-12) show that we perform better that all baselines. All results are comparable to the State-of-the-Art on TUM RGB-D except for PWC-Net, likely due to its naive integration (if there is no object moving the predicted mask will be wrong). However, on the Consensus Inversion dataset results are quite different: Mask R-CNN and PWC-Net both perform poorly while RVOS and COSNet have very good results. This shows that methods that are class-agnostic perform better than class-aware methods and -- considering the other columns of the table -- geometrical methods. The main issue with UVOS networks is in fact oversegmentation: RVOS and COSNet failed on the \emph{Static} sequences of the Consensus Inversion / Static subset. They masked the only source of features in the image and made the SLAM fail. The same conclusions come from Table \ref{tab:success_sota}. Fig. \ref{fig:uvos} illustrates failure cases.

The conclusion is that UVOS networks are better at making SLAMs robust to dynamic objects than the usual semantic and geometrical approaches. However, they also segment static objects and have an increased risk of failing early in static environments. 

\subsubsection{Evaluation of monocular false starts} monocular SLAMs as ORB-SLAM 2 require the camera to move to initialize, so we evaluate a specific kind of failure: false starts. Fig. \ref{fig:static_cam} illustrates such a false start: since the camera is static, generated maps or trajectories are necessarily fake. We evaluate the State-of-the-Art and the baselines on the Consensus Inversion / Static subset. We performed best, never initializing incorrectly. All other methods failed, either because the object is unknown (semantic approaches), because the object caused a consensus inversion (geometrical approaches) or because it was not fully segmented (UVOS approaches). The results show again that it is essential to make a SLAM robust in a specific environment.

\subsubsection{Extension to a Direct SLAM}: we test LDSO, a monocular direct SLAM. We improve it by integrating the network we trained using ORB-SLAM 2 after the feature detector. Fig. \ref{fig:ldso_inversion} shows the effect of masking: the 3D reconstruction is incorrect if the object is not masked. However, when masked, the SLAM operates normally. Table \ref{tab:ate_ldso} shows that both the Penalized ATE RMSE and Success Rate improve. While the Success Rate does not reach 100\%, the result is very interesting: it is possible to use one SLAM to improve another one.

\begin{table}
    \centering
    \begin{adjustbox}{max width=0.49\textwidth}
    \begin{tabular}{|c|c|c|}
        \cline{2-3}
        \multicolumn{1}{c|}{} & \multicolumn{2}{|c|}{\textbf{LDSO} \cite{gao_ldso_2018} + $\cdots$} \\ 
        \multicolumn{1}{c|}{} &  \textbf{No segmentation} & \textbf{Our segmentation} \\ \hline
        Avg. Penalized ATE RMSE (m) & 0.0833 & \textbf{0.0581} \\ \hline
        Success Rate (\%) & 36.4\% & \textbf{63.6\%} \\ \hline
    \end{tabular}
    \end{adjustbox}
    \vspace*{0mm}
    \caption{Avg. Penalized ATE RMSE (m) and Success Rate (\%) of LDSO and our masked version on the Consensus Inversion/Dynamic dataset.}
    \label{tab:ate_ldso}
\end{table}

\subsection{Limitations}
There are some limitations to our method. We rely on video segmentation networks but they may not work in ambiguous situations, e.g., when a dynamic object passes in front of a very similar object. Dynamic objects must be reconstructed by the SLAM to generate outliers so it is difficult to mask objects that move all the time and we do not yet handle new dynamic objects at runtime. Evidently, the improved SLAM stops tracking if a known dynamic object covers the image.

\section{Conclusions}
In this paper we proposed a novel method to learn to segment dynamic objects using only one monocular sequence per dynamic object, which is an advantage in comparison to previous methods. More importantly, we do not need any manual labelling which makes our method much easier to use. 

 We also presented the Consensus Inversion dataset and new metrics to evaluate the robustness of Dynamic SLAMs. We showed that consensus inversions can cause major SLAM failures, even to State-of-the-Art Dynamic SLAMs. We improved ORB-SLAM 2 monocular/stereo/RGB-D as well as LDSO and achieved top results in very challenging scenarios. 
 
 Another advantage, in addition to preventing SLAM failures and improving motion estimation, is the improvement in map reconstruction. Tasks like relocalization and loop closing need accurate maps and should benefit from our approach.

\section{Acknowledgements} This publication was made possible by the use of the FactoryIA supercomputer, financially supported by the Ile-de-France Regional Council.

\bibliographystyle{IEEEtran}
{
\bibliography{egbib}}

\end{document}